\documentclass[letterpaper]{article} % DO NOT CHANGE THIS
\usepackage{aaai2026}  % DO NOT CHANGE THIS
\usepackage{times}  % DO NOT CHANGE THIS
\usepackage{helvet}  % DO NOT CHANGE THIS
\usepackage{courier}  % DO NOT CHANGE THIS
\usepackage[hyphens]{url}  % DO NOT CHANGE THIS
\usepackage{graphicx} % DO NOT CHANGE THIS
\usepackage{natbib}  % DO NOT CHANGE THIS AND DO NOT ADD ANY OPTIONS TO IT
\usepackage{caption} % DO NOT CHANGE THIS AND DO NOT ADD ANY OPTIONS TO IT
\usepackage{algorithm}
\usepackage{algorithmic}

\usepackage{newfloat}
\usepackage{listings}
\DeclareCaptionStyle{ruled}{labelfont=normalfont,labelsep=colon,strut=off} % DO NOT CHANGE THIS
\floatstyle{ruled}
\newfloat{listing}{tb}{lst}{}
\floatname{listing}{Listing}
\usepackage{mdframed}  
\usepackage{fancyvrb}
\usepackage{amsmath}
\usepackage{booktabs} 
\usepackage{fvextra}

\title{From Policy to Logic for Efficient and Interpretable Coverage Assessment}
\author{
    Rhitabrat Pokharel\textsuperscript{\rm 1},
    Hamid Reza Hassanzadeh\textsuperscript{\rm 2},
    Ameeta Agrawal\textsuperscript{\rm 1}
}
\affiliations{
    \textsuperscript{\rm 1}Department of Computer Science, Portland State University\\
    \textsuperscript{\rm 2}Optum AI\\
    \{pokharel, ameeta\}@pdx.edu, hamid.hassanzadeh@optum.com
}

\usepackage{bibentry}
\begin{document}

\maketitle

\begin{abstract}
Large Language Models (LLMs) have demonstrated strong capabilities in interpreting lengthy, complex legal and policy language. However, their reliability can be undermined by hallucinations and inconsistencies, particularly when analyzing subjective and nuanced documents. These challenges are especially critical in medical coverage policy review, where human experts must be able to rely on accurate information. In this paper, we present an approach designed to support human reviewers by making policy interpretation more efficient and interpretable. We introduce a methodology that pairs a coverage-aware retriever with symbolic rule-based reasoning to surface relevant policy language, organize it into explicit facts and rules, and generate auditable rationales. This hybrid system minimizes the number of LLM inferences required which reduces overall model cost. Notably, our approach achieves a 44\% reduction in inference cost alongside a 4.5\% improvement in F1 score, demonstrating both efficiency and effectiveness.
\end{abstract}


\section{Introduction}
\begin{figure}[!t]  
    \centering  
    \includegraphics[width=0.27\textwidth]{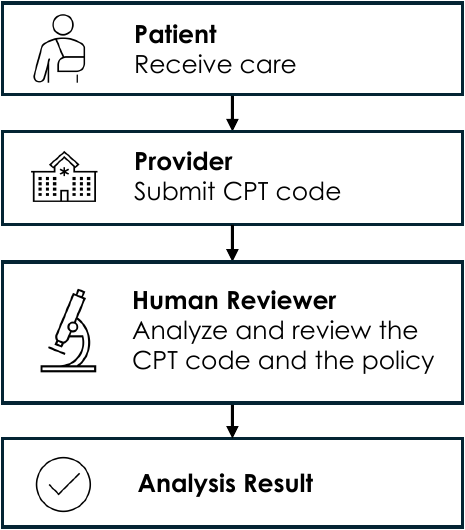}
    \caption{High level pipeline of CPT code analysis in healthcare.}  
    \label{fig:cpt_spd_pipeline}
\end{figure}

Healthcare procedures encompass a wide range of diagnostic, therapeutic, and preventive services. From routine check-ups and laboratory tests to complex surgical interventions, each procedure must be accurately recorded and communicated across healthcare systems to ensure quality care and regulatory compliance. To achieve this standardization, the healthcare industry relies on codes like Current Procedural Terminology (CPT) codes. Healthcare providers, insurance companies, and other stakeholders rely on these codes to communicate effectively and process claims for a wide range of medical procedures. Accurate interpretation of CPT codes and policy documents is essential for determining whether the code aligns with specific policy provisions. Figure~\ref{fig:cpt_spd_pipeline} illustrates a high level workflow involved in CPT code analysis. Given the intricate and often subjective nature of healthcare policy documents, this process can be time-consuming and susceptible to inconsistencies. 
These challenges highlight the need for more efficient and reliable approaches to support policy logic tracing and ensure consistency within the healthcare domain. 
% These challenges underscore the need for tools that enhance efficiency, improve interpretability, and support human reviewers in making consistent, well-grounded determinations within the healthcare domain.

Recent advances in Large Language Models (LLMs) have demonstrated considerable promise in decision-making across domains such as legal analysis and healthcare policy interpretation \citep{ryu2025divide, pan-etal-2023-logic, xu-etal-2024-faithful}, owing to their ability to process and interpret complex natural language. While it is generally accepted that LLMs are adept at analyzing textual information, they often face difficulties when analyzing  lengthy complex texts. Effective reasoning in these contexts requires models to explicitly reference factual details and policy language, which are frequently dispersed throughout the document. Furthermore, LLMs may exhibit hallucinations and inconsistencies in their reasoning \citep{LegalfictionsDahletal2024}.

Chain-of-thought (CoT) prompting \citep{wei2022chain} is a commonly explored technique \citep{ kant2025towards} for guiding LLMs through multi-step reasoning processes. However, CoT approaches are not immune to lack of interpretability, generating inconsistent reasoning, and can be computationally expensive when applied at scale. 

% Chain-of-thought (CoT) prompting \citep{wei2022chain} with LLMs is a commonly explored approach for reasoning over natural language \citep{ kant2025towards}. While this method can effectively guide the model through multi-step reasoning processes, it is prone to hallucinations, where the model generates information that is not grounded in the contract, and inconsistencies in its reasoning. Additionally, relying on CoT to interpret a CPT code each time a decision is to be made can be computationally expensive, making it less practical for large-scale applications. These challenges emphasize the need for more robust and cost-efficient solutions.

Expert systems have been used to simulate the reasoning abilities of human experts by encoding domain-specific knowledge and reasoning processes \citep{li2024mediq, garrido2025gofai}. These systems operate on structured representations of knowledge, such as facts and ``if-then" rules, using deterministic logic to systematically process knowledge. By applying predefined rules, it enables expert systems ensure consistency and interpretability. 

% In this paper, we introduce a neuro-symbolic approach that simulates the decision-making capabilities of expert systems to assist human reviewers with decision tracing. Our method combines the strengths of neural networks for processing natural language with symbolic reasoning to apply structured rules. By converting policy terms into a machine-interpretable format, this approach enables logical reasoning akin to that of expert systems. As such, this system presents interpretable rationales for each coverage decision, transparently tracing its reasoning process back to the source policy document. Importantly, the human reviewer retains final adjudication authority, with our method serving as an assistant that supports auditable reasoning.

In this paper, we introduce a neuro-symbolic approach that emulates the structured reasoning patterns of expert systems to assist human reviewers with understanding and tracing policy logic. Our method integrates neural components for processing natural language with symbolic reasoning modules that apply clearly defined rules. By translating policy terms into a machine-interpretable format, the system can organize and surface the relevant logical conditions underlying a policy. In doing so, it provides interpretable rationales that point directly back to the governing policy language. Crucially, the system does not make coverage determinations; human reviewers maintain full adjudication authority. Instead, our approach serves as a tool to find support from coverage documents to support auditable reasoning throughout the review process.

Our major contributions are as following.

\begin{itemize}
    \item We develop a framework that supports human experts with analyzing complex documents by integrating neuro-symbolic approach.
    \item To support transparent tracing, we implement a coverage-aware retriever that accurately identifies and extracts governing policy language relevant to specific CPT codes.
    \item We provide a rule-based system that significantly lowers the cost of reasoning compared to continuous LLM inference.
\end{itemize}

% \begin{figure*}[t]
% \small
% % \scriptsize
% \footnotesize
% \begin{mdframed}
% % \begin{Verbatim}
% \begin{Verbatim}[breaklines=true, breakanywhere=true, breaksymbol={}]
% Diabetes SelfManagement and Training/Diabetic Eye Exams/Foot Care Outpatient self-management training for the treatment of diabetes, education and medical nutrition therapyservices. Services must be ordered by a Physician and provided by appropriately licensedor registered health care professionals who are authorized to prescribe such items and whodemonstrate adherence to minimum standards of care for diabetes mellitus as adopted andpublished by the Diabetes Initiative. Benefits also include medical eye exams (dilatedretinal exams) and preventive foot care for diabetes. Diabetic Self Management ItemsInsulin pumps and supplies and continuous glucose monitors for the management andtreatment of diabetes, based upon your medical needs. An insulin pump is subject to allthe conditions of coverage stated under Durable Medical Equipment (DME), Orthotics andSupplies. Benefits for blood glucose meters including continuous glucose monitors, insulinsyringes with needles, blood glucose and urine test strips, ketone test strips and tabletsand lancets and lancet devices are described under the Outpatient Prescription Drug Rider.
% \end{Verbatim}
% \end{mdframed}
% \caption{A sample text from \textit{Diabetes Services} subsection of a plan document.}
% \label{fig:sample_subsection}
% \end{figure*}

\section{Related Work}
\subsection{Translating Natural Language into First-Order Logic}

The use of LLMs to translate natural language into formal rules has gained significant attention in recent research. Techniques such as prompting with CoT and neuro-symbolic approaches \cite{wei2022chain, nezhad2025symcode, pmlr-v284-nezhad25a} have shown promising results in this area. Contractual language often encodes logical relationships in natural language that humans interpret with ease. Several recent systems, including CLOVER \citep{ryu2025divide}, LOGIC-LM \citep{pan-etal-2023-logic}, LogicLLaMA \citep{yang-etal-2024-harnessing}, ProSLM \citep{vakharia2024proslm}, Thought Like Pro \citep{tan2024thought}, SymbCoT \citep{xu-etal-2024-faithful}, and LLM-Tres \citep{toroghi-etal-2024-verifiable}, have explored methods for translating natural language into first-order logic representations. These approaches aim to bridge the gap between unstructured contract language and machine-interpretable rules. Our work is in line with these efforts to formalize natural language reasoning, but differs in that we employ a rule-based expert system to operationalize domain knowledge through executable reasoning rather than formal logical translation.

\subsection{LLMs for Legal Reasoning}
LLMs have demonstrated notable progress in supporting legal decision-making tasks, including the following examples. \citet{guha2023legalbench} introduced LegalBench, a comprehensive suite of benchmarks to evaluate the legal reasoning capabilities of LLMs. \citet{yao-etal-2025-elevating} proposed a reinforcement learning-based approach for legal question answering. \citet{mishra-etal-2025-investigating} conducted an error analysis of LLM reasoning in civil procedure contexts. \citet{LegalfictionsDahletal2024} examined the prevalence of hallucinations in LLMs applied to legal domains. \citet{Shengbin2024lawllm} presented LawLLM, a system offering legal reasoning services through fine-tuning techniques. \citet{shen2025law} introduced a reasoning schema for legal tasks that integrates factual grounding. Similarly, \citet{shi2025legalreasoner} proposed LegalReasoner, which first identifies disputes to decompose complex cases and then performs step-wise reasoning. Our approach focuses on using actionable rules generated from text.

% In the field of decision making in Law, LLMs have shown significant achievements. \cite{guha2023legalbench} release suite of benchmark for legal reasoning ability of LLMs. \cite{yao-etal-2025-elevating} use reinforcement learning based approach for legal QA. 
% \cite{mishra-etal-2025-investigating} studied error analysis in LLM reasoning in civil procedures.
% \cite{LegalfictionsDahletal2024} talked about prevalent hallucinations in LLMs in legal domain.
% \cite{Shengbin2024lawllm} propose this LawLLM system that provide legal reasoning services using finetuning method.
% \cite{shen2025law} provide with reasoning schema for legal reasoning that is supported by facts.
% \cite{shi2025legalreasoner} first identifies disputes to decompose complex cases and then does step wise reasoning. 

\begin{figure*}[t]
    \centering  
    \includegraphics[width=1.0\textwidth]{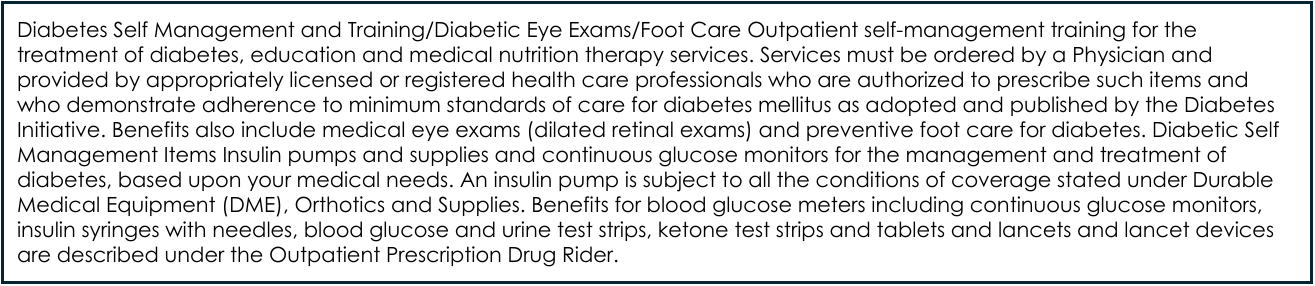}
    \caption{A sample text from \textit{Diabetes Services} subsection of a plan document.}  
    \label{fig:sample_subsection}
\end{figure*}

\begin{figure*}[!t]
    \centering  
    \includegraphics[width=1.0\textwidth]{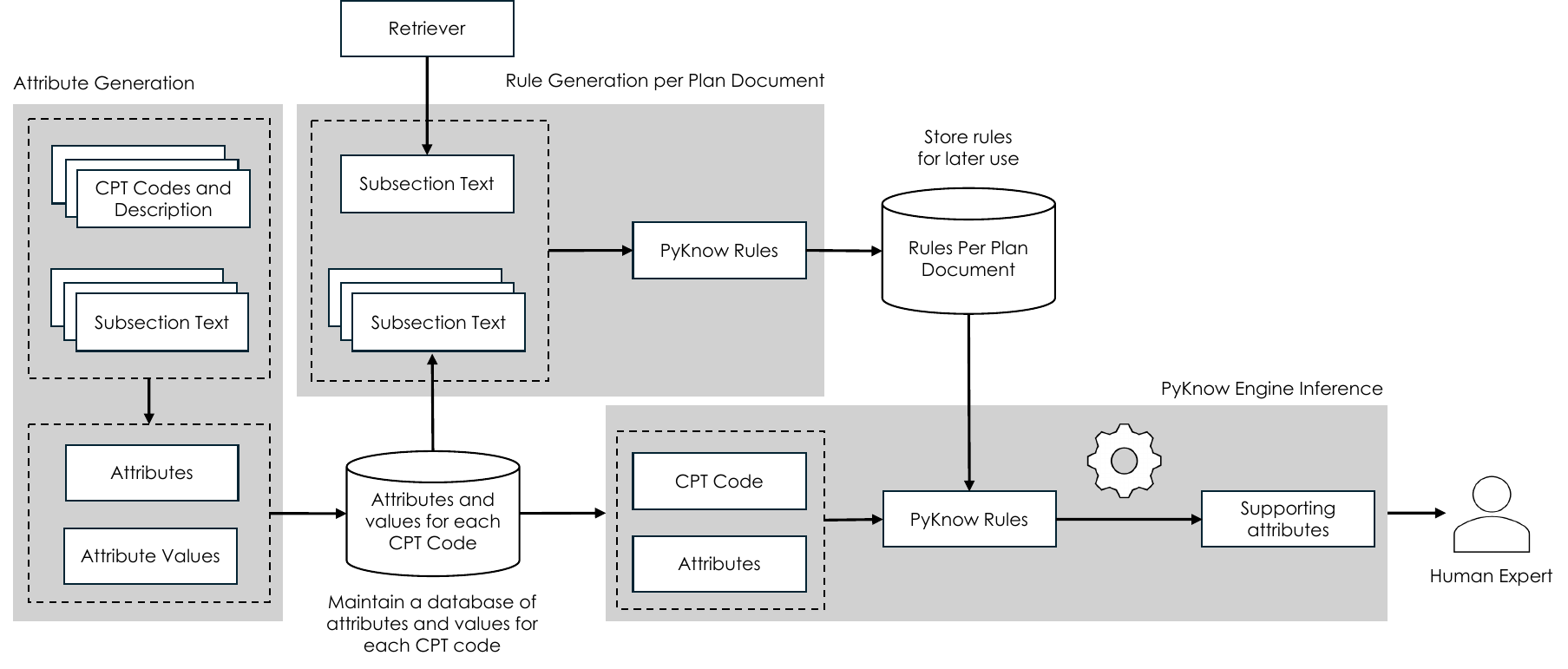}
    \caption{Pipeline of our neuro-symbolic approach. This system supports human reviewers by making the underlying policy logic interpretable.}
    \label{fig:rule_workflow}
\end{figure*}

\subsection{LLMs for Rule-Making}

% \cite{cummins2025computable} explored the potential of using computable rules for a specific domain using Prolog-like logical representations.  \cite{kant2025towards} conducted an extensive evaluation of current LLMs in producing structured rules. In their study, they analyze a policy document by comparing the decision-making ability of LLMs using a prompt-based approach versus a rule-based approach. Here, LLMs are employed to generate structured rules. The results demonstrate that most LLMs show enhanced reasoning performance when integrated with rule-based frameworks. Their method uses human-designed schemas with specific attributes and helper functions to guide the generation of structured rules. The approach is constrained to reasoning within the bounds of the provided facts. In contrast, our work eliminates the need for human-designed schemas and helper functions, focusing on dynamic rule generation directly from natural language, while enabling reasoning that extends beyond the explicitly stated information.

\citet{cummins2025computable} explored the potential of using computable rules for decision-making in a specific domain, leveraging Prolog-like logical representations to encode policy logic in a machine-readable format. Their work highlights the importance of formal rule structures in enabling automated reasoning, but relies heavily on manual encoding and domain expertise for rule construction.

Building on this foundation, \citet{kant2025towards} conducted an extensive evaluation of the current generation of LLMs in producing structured rules from policy documents. In their study, the authors systematically compared the decision-making capabilities of LLMs using both prompt-based approaches and rule-based frameworks. LLMs were tasked with generating structured rules from policy text, guided by human-designed schemas featuring specific attributes and helper functions. Their findings show that most LLMs exhibit markedly improved reasoning performance when integrated with explicit rule-based scaffolding, underscoring the value of structured logic in complex adjudication tasks. However, their approach is limited by the scope of the provided schemas and helper functions, constraining reasoning to facts explicitly represented in the input.

In contrast to prior approaches, our work advances this line of research by eliminating the need for human-curated schemas and helper functions. We focus on dynamic rule generation from natural language, leveraging finetuned models and symbolic reasoning to automatically extract governing policy language and generate rules with the intent to offers a scalable and cost-effective solution that reduces manual effort and reliance on frequent LLM inference.

\section{Coverage Documents}

Coverage documents serve as references that outline the criteria and provisions associated with a given policy. They are organized into multiple sections and subsections, each providing detailed information about the procedures, services, or conditions addressed by the policy. These subsections contain specific language that clarifies the scope of the policy that guides how particular provisions apply in various contexts. A sample subsection is presented in Figure~\ref{fig:sample_subsection}.

% Coverage documents serve as references that outline the criteria for determining whether specific procedures or services are covered under a given policy. These documents are typically structured into two main sections: inclusion and exclusion. The inclusion section generally details the procedures, services, or conditions that are covered, while the exclusion section specifies those that are not covered; however, it is important to note that not everything listed under inclusion is always covered, nor is everything under exclusion categorically not covered, as there may be specific conditions or criteria that determine inclusion or exclusion in both sections. Each of these sections is further divided into subsections that provide detailed explanations of the covered or non-covered items. These subsections provide detailed guidance, helping to clarify the extent of coverage and support accurate decision-making in accordance with policy provisions. A sample subsection is presented in Figure~\ref{fig:sample_subsection}.

\section{Methodology}
\label{sec:method}

We define our task as follows: given a CPT code, its accompanying description, and a policy document, the objective is to generate an interpretable reasoning trace that links the code to the relevant policy language. We first describe our approach for extracting relevant coverage language from policy documents, and then discuss the process of formulating symbolic rules. Overall pipeline is shown in Figure~\ref{fig:rule_workflow}.

\subsection{Policy Text Retrieval Phase} \label{subsec:retriever}
A core challenge in reasoning over symbolic rules is retrieving policy language that governs coverage, rather than simply matching on topical similarity. Standard semantic search is ill-suited for this task because the signals that determine benefit status are often orthogonal to thematic content. For instance:

\begin{itemize}
    \item A CPT for insulin pump initiation is thematically close to passages about diabetes self-management education, nutrition therapy, or endocrinology follow-ups, none of which determine its benefit status. The governing rule is more likely a short paragraph under Durable Medical Equipment or a specific policy rider.
    \item A CPT for continuous glucose monitoring may cluster with general diabetes monitoring advice or HbA1c screening policies, while the actual coverage clause is often a concise exclusion or a limitation found in a separate section.
    \item CPTs for debridement or wound care sit near content about diabetic foot care or peripheral neuropathy, whereas the decisive language is typically found in surgical necessity sections that include prior-authorization requirements.
\end{itemize}

To align retrieval with what matters for reasoning, we developed a coverage-aware retriever. Instead of optimizing for semantic likeness, we trained a cross-encoder to score subsections by their likelihood of explicitly governing a CPT’s coverage, limitations, or exclusions.

\noindent \textbf{Expert-Labeled Supervision.} We built an internal annotation platform and engaged approximately 20 certified coding Subject Matter Experts (SMEs), with an arbitration process to ensure consistency. For each CPT across 172 Certificates of Coverage/Summary Plan Documents (CoCs/SPDs), these SMEs selected only the passages that decide coverage. This effort produced over 1.84 million labeled (CPT, subsection, relevance) pairs, with 10\% reserved for validation and the remainder ($\sim1.61$ million) used for training. This dataset provides the direct supervision needed to learn the nuances of policy language.  
  
\noindent \textbf{Formulation as Contrastive Multiple-Choice Ranking.} The task is framed as a multiple-choice ranking problem with a contrastive objective. For a given CPT query $q$ and its set of candidate subsections from a plan, $\{s_1, \dots, s_n\}$, one passage $s_i$ is labeled positive (the true coverage passage), and the rest are negatives. The model processes each query-passage pair $(q, s_i)$ to output a logit; a softmax over all logits yields a probability distribution. Training minimizes the cross-entropy loss on the positive label:  
  
$  
\\
\mathcal{L} = -\log p(i = \text{positive} \mid q, S)  
\\
$  
  
This formulation, functionally equivalent to contrastive objectives like InfoNCE \citep{oord2018representation}, forces the model to distinguish the single governing passage from a set of highly relevant but non-dispositive distractors from the same document. Queries are constructed as:  
  
$ 
\\ 
\langle \text{CPT} \rangle : \langle \text{lay description} \rangle  
\\
$  
  
and choices are the raw subsection texts.  

\noindent \textbf{Architecture and Training.}
The model is a LongformerForMultipleChoice fine-tuned from the allenai/longformer-base-4096\footnote{https://huggingface.co/allenai/longformer-base-4096} backbone. Its 1,536-token context window is sufficient to process entire subsections, including nested formatting, without truncation. Key training parameters include the AdamW optimizer (learning rate 2e-5, weight decay 0.01), bf16 mixed precision, and gradient checkpointing for memory efficiency. The model was trained for 2.5 epochs ($\sim$48 hours) on a single node with 8 x H100 GPUs, with periodic evaluation and checkpointing. To accelerate experimentation, the dataset was pre-tokenized once and reused across training runs.

\noindent \textbf{Why a Cross-Encoder Is Feasible and Effective.} A cross-encoder architecture, which jointly processes the query and passage through its attention layers, is essential for this task. It can capture the fine-grained interactions between a CPT code and subtle policy phrases, such as ``prior authorization required," ``not covered," or ``limited to", which are often lost in compressed vector representations. While computationally intensive, this approach is feasible in our setting because the candidate pool per plan is small and well-defined (typically $<$ 60 subsections across the relevant ``Covered Services" and ``Exclusions \& Limitations" sections). Exhaustive scoring requires a trivial number of forward passes per CPT on modern GPUs, and the gains in precision are substantial. The model learns to prioritize short, auditable policy fragments that a downstream rule-based engine can deterministically parse, making the entire RAG stack coverage-aware by design.

\noindent \textbf{Inference Pipeline.}
\begin{itemize}
    \item Preparation: For each project, all subsections from the relevant plan sections are loaded as the candidate pool.
    \item Scoring: For each CPT, the query is constructed and scored against every candidate subsection using the trained cross-encoder. Logits are softmax-normalized across the candidate set for that CPT.
    \item Filtering: Passages with a probability above a threshold $\tau$ (default 0.25) are retained, capped at a maximum of five from ``Covered Services" and five from ``Exclusions \& Limitations." In real-world deployment, if retrieval returns no passage above $\tau$, the system should escalate the case for human review.
    \item Output: Results are streamed progressively, with each row containing project id, CPT code, probability, section, subsection text, and other metadata. If no passage clears the threshold for a CPT, a placeholder row is emitted to ensure downstream stages can track completeness.
\end{itemize}

\subsection{Symbolic Phase}

In this phase, a rule-based system is created to systematically encode the coverage criteria to help process the CPT codes. This system is implemented using PyKnow\footnote{https://github.com/buguroo/pyknow}, a Python library for symbolic reasoning. PyKnow facilitates the creation of expert systems by offering tools to define facts (units of information), fields (data within facts), and rules (logic for reasoning over facts). In this study, we use the term ``attributes" to refer to the fields. Our methodology first extracts attributes associated with each CPT code, followed by generating rules derived from coverage text, and finally executing the PyKnow inference engine to apply these rules and simulate expert reasoning over the data.

% \noindent \textbf{CPT Code-to-Subsection Mapping.} Mapping each CPT code to its possible subsections from the respective coverage document is a critical step in this approach. This process is essential not only for attribute generation but also during inference. The rules governing this step are often complex and lengthy. By identifying the possible subsections associated with a given CPT code, decisions can be made within these specific subsections rather than relying on rules applicable to all subsections. This approach significantly reduces the time required during inference, while simultaneously aiding the generation of relevant attributes for the CPT code. This matching uses both zero-shot and finetuned method separately from retriever step.

% The utilization of subsections in conjunction with CPT codes is discussed below. 

\begin{figure*}[!t]
    \centering  
    \includegraphics[width=0.95\textwidth]{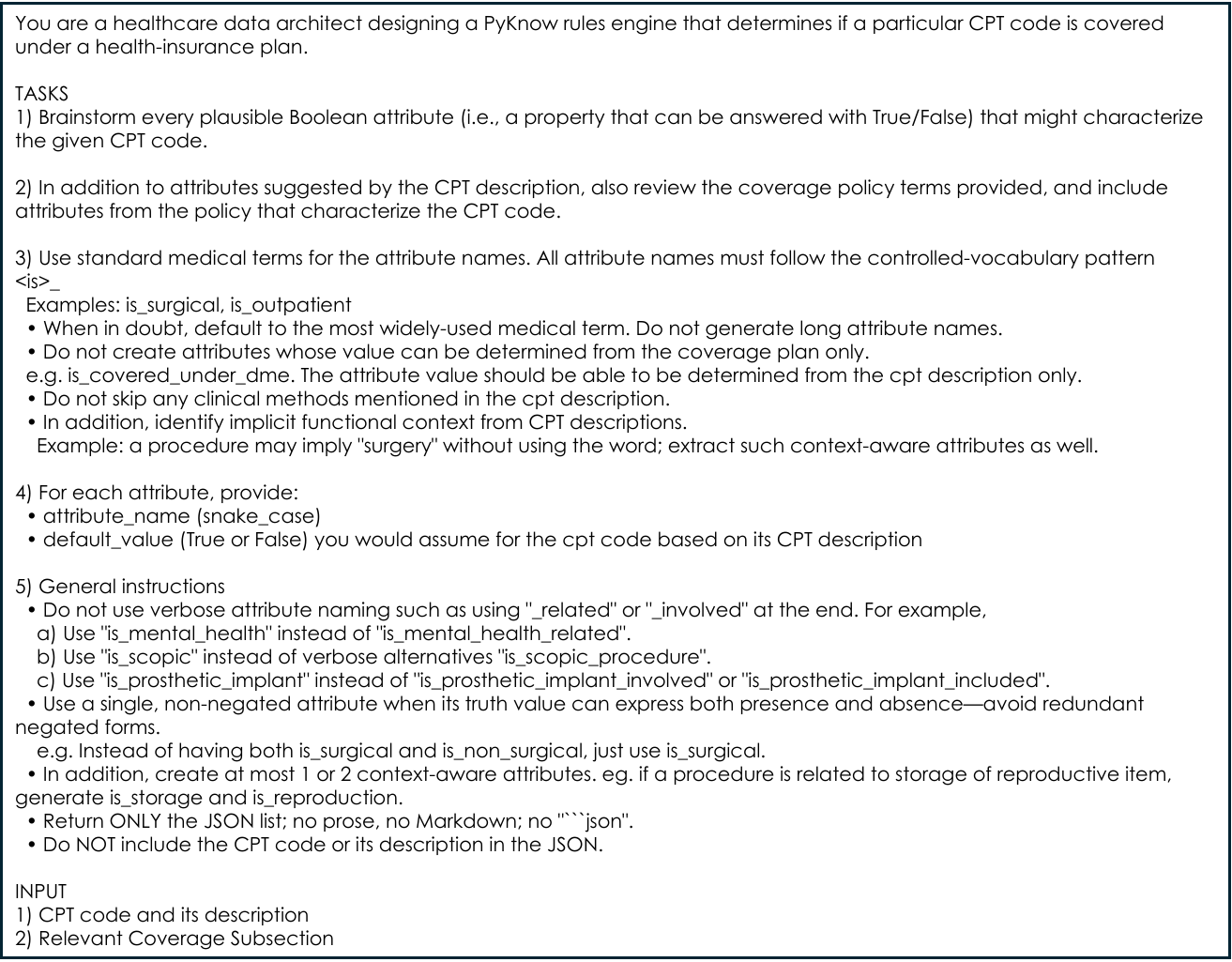}
    \caption{The prompt used for attribute generation.}  
    \label{fig:prompt_attr}
\end{figure*}

\begin{figure}[t]
    \centering  
    \includegraphics[width=0.37\textwidth]{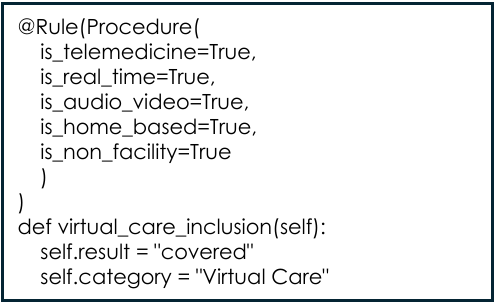}
    \caption{An example of a rule generated. Each rule is in this similar format.}
    \label{fig:sample_rule}
\end{figure}

\begin{table*}[t]
\centering
\begin{tabular}{lcccccc}
\toprule
\textbf{Plan} & \multicolumn{3}{c}{\textbf{Accuracy}} & \multicolumn{3}{c}{\textbf{F1 Score}} \\
\cmidrule(lr){2-4} \cmidrule(lr){5-7}
 & \textbf{GPT4.1} & \textbf{Rule-based (ZR)} & \textbf{Rule-based (FR)} & \textbf{GPT4.1} & \textbf{Rule-based (ZR)} & \textbf{Rule-based (FR)} \\
\midrule
Plan \#1 & 0.88 & 0.87 & 0.81 & 0.93 & 0.93 & 0.90 \\
Plan \#2 & 0.78 & 0.84 & 0.82 & 0.87 & 0.91 & 0.90 \\
Plan \#3 & 0.77 & 0.79 & 0.89 & 0.86 & 0.88 & 0.94 \\
Plan \#4 & 0.88 & 0.83 & 0.88 & 0.93 & 0.90 & 0.93 \\
Plan \#5 & 0.83 & 0.90 & 0.92 & 0.90 & 0.94 & 0.96 \\
Plan \#6 & 0.77 & 0.83 & 0.88 & 0.86 & 0.90 & 0.93 \\
Plan \#7 & 0.88 & 0.88 & 0.90 & 0.93 & 0.93 & 0.94 \\
\midrule
\textbf{Average} & 0.82 & 0.85 & \textbf{0.87} & 0.89 & 0.91 & \textbf{0.93} \\
\bottomrule
\end{tabular}
\caption{Accuracy and F1 scores per plan. ZR = Zero-shot Retriever; FR = Finetuned Retriever. Overall, the finetuned rule-based system outperforms the other methods.}
\label{tab:results}
\end{table*}

\noindent \textbf{Attribute Generation.} Attributes represent distinct properties that characterize a CPT code and relate it to specific policy provisions. Each attribute reflects a specific characteristic of the procedure, such as whether it pertains to mental health or preventive care. For a given coverage document, once each CPT code is mapped to its relevant subsections (in the retrieval phase), the codes are grouped by subsection. To generate attributes for each CPT code, the model is prompted to identify the properties that describe the CPT code and are likely shared with the previously grouped subsections. To facililate this, a short description of the code is also provided. Simultaneously, the model assigns default values (True or False) to these attributes based on the extracted information. Each attribute is framed as a yes/no question, meaning it should provide a clear ``yes" or ``no" response regarding the applicability of the attribute to the CPT code. For example, an attribute might be \texttt{is\_implant}, where the value is either \texttt{True} or \texttt{False}. The prompt used for this process is detailed in Figure~\ref{fig:prompt_attr}. Attributes for CPT codes are created only once, and the same attributes can be reused for new plan documents. This approach ensures scalability while minimizing costs. We use 10 plan documents during this step.

\noindent \textbf{Rule Creation.} In this step, symbolic rules are generated using a well-structured and guided prompt based on the coverage text and relevant attributes. For each subsection, the associated CPT codes are grouped along with their attributes. These attributes help represent the coverage document in the form of PyKnow rules. The reason for incorporating attributes during rule creation is to ensure that rules are created using only the relevant attributes. Without this, attributes not associated with the CPT codes might be created, leading to potential syntax errors. For each plan document, a distinct set of rules is generated, with rules created for each subsection within the document. The prompt used during this process is in Appendix~\ref{apd:prompts}. Specific instructions are provided to ensure that the generated rules are free from syntax errors. A sample rule is presented in Figure~\ref{fig:sample_rule}.

\noindent \textbf{Inference.} In this step, given a CPT code and its associated attributes, we use the PyKnow engine to identify which rule is triggered. After a rule is matched, the relevant attributes are passed on to the human reviewer for further analysis.
% we trace the decision back to the relevant attributes, which we describe further in the results section. 
% This step does not require a costly hardware like GPU.

% \begin{figure}[t]
% \small
% \begin{mdframed}
% \begin{Verbatim}
% @Rule(Procedure(
%     is_telemedicine=True,
%     is_real_time=True,
%     is_audio_video=True,
%     is_home_based=True,
%     is_non_facility=True
%     )
% )
% def virtual_care_inclusion(self):
%     self.result = "covered"
%     self.category = "Virtual Care"
% \end{Verbatim}
% \end{mdframed}
% \caption{Each rule are in this similar format.}
% \label{fig:sample_rule}
% \end{figure}

\section{Experimental Settings}

% avoiding the use of CPT code (since it is licensed)
% \subsection{Dataset}
We use internal data, which consists of coverage documents, procedure descriptions, and corresponding human-generated determinations. The coverage documents are in the form of a plain text. To ensure compliance with privacy policies, both patient information and details related to specific plan documents are fully anonymized. The evaluation dataset consists of the same 814 CPT codes across 7 separate coverage documents (i.e. total of 5,698 codes) that were never used in the retriever training or attribute creation phase. The same dataset is used to get all the results.

% \subsection{Models}
For our baseline experiment, we use GPT-4.1 baseline with vanilla prompting along with CPT code and the entire plan document. The rule-based approach also utilizes GPT-4.1 for initial processing (attribute generation and rule-making). Additionally, we include other models such as o3 and GPT5-mini in an ablation study. For evaluation, we report accuracy and F1 scores.

% Reasoning models, such as o3, are not used for two primary reasons: first, our objective is to avoid relying solely on the reasoning capabilities of such models; second, reasoning models are cost-prohibitive, which contradicts our goal of reducing overall expenses. 

\section{Results}

\begin{figure*}[t]
    \centering  
    \includegraphics[width=0.7\textwidth]{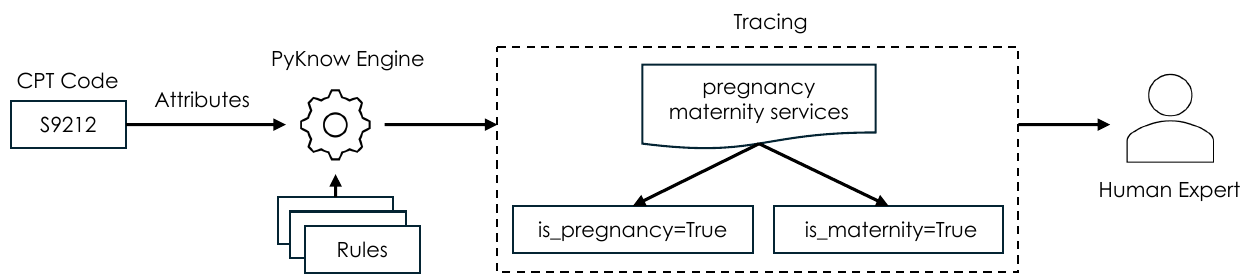}
    \caption{Overall inference workflow illustrating how our system supports human experts in identifying relevant support from coverage documents.}  
    \label{fig:rule_tracing}
\end{figure*}

% \begin{figure}[!t]
%     \centering  
%     \includegraphics[width=0.5\textwidth]{images/acc_rule.png}
%     \caption{Average accuracy scores per plan.}  
%     \label{fig:acc_rule}
% \end{figure}

% \begin{figure}[!t]
%     \centering  
%     \includegraphics[width=0.5\textwidth]{images/f1_rule.png}
%     \caption{F1 accuracy scores per plan.}  
%     \label{fig:f1_rule}
% \end{figure}

% \begin{table}
% \centering
% \begin{tabular}{lccc}
% \toprule
% \textbf{Plan} & \textbf{GPT4.1} & \textbf{Rule (ZR)} & \textbf{Symbolic (FR)} \\
% \midrule
% Plan \#1 & 0.88 & 0.87 & 0.81 \\
% Plan \#2 & 0.78 & 0.84 & 0.82 \\
% Plan \#3 & 0.77 & 0.79 & 0.89 \\
% Plan \#4 & 0.88 & 0.83 & 0.88 \\
% Plan \#5 & 0.83 & 0.90 & 0.92 \\
% Plan \#6 & 0.77 & 0.83 & 0.88 \\
% Plan \#7 & 0.88 & 0.88 & 0.90 \\
% \midrule
% Average & 0.82 & 0.85 & \textbf{0.87} \\
% \bottomrule
% \end{tabular}
% \caption{Average accuracy scores per plan. ZR = Zero-shot Retriever; FR = Finetuned Retriever.}
% \label{tab:avg_acc}
% \end{table}

% \begin{table}
% \centering
% \begin{tabular}{lccc}
% \toprule
% \textbf{Plan} & \textbf{GPT4.1} & \textbf{Rule (ZR)} & \textbf{Symbolic (FR)} \\
% \midrule
% Plan \#1 & 0.93 & 0.93 & 0.90 \\
% Plan \#2 & 0.87 & 0.91 & 0.90 \\
% Plan \#3 & 0.86 & 0.88 & 0.94 \\
% Plan \#4 & 0.93 & 0.90 & 0.93 \\
% Plan \#5 & 0.90 & 0.94 & 0.96 \\
% Plan \#6 & 0.86 & 0.90 & 0.93 \\
% Plan \#7 & 0.93 & 0.93 & 0.94 \\
% \midrule
% Average & 0.89 & 0.91 & \textbf{0.93} \\
% \bottomrule
% \end{tabular}
% \caption{Average F1 scores per plan.}
% \label{tab:avg_f1}
% \end{table}

\begin{table*}[ht]  
\centering  
\begin{tabular}{lccccc}  
\toprule  
\textbf{Model} & \textbf{Context Provided} & \textbf{Acc.} & \textbf{F1} & \textbf{Cost per 1k CPTs} & \textbf{Cost for 11k CPTs} \\  
\midrule  
GPT-5-mini (FR)             & Retrieved text     & 0.94 & 0.96 & \$440    & \$4,840 \\  
GPT-4.1 (FR)                 & Retrieved text     & 0.92 & 0.95 & \$880    & \$9,680 \\  
O3 (FR)                      & Retrieved text     & 0.94 & 0.96 & \$880    & \$9,680 \\  
GPT-4.1                 & Entire document    & 0.82 & 0.89 & \$3,520  & \$38,720 \\ 
\midrule
Rule-based (ZR)         & Retrieved text     & 0.85 & 0.91 & \$2   & \$22 \\  
Rule-based (FR)         & Retrieved text     & 0.87 & 0.93 & \$2   & \$22 \\  
\bottomrule  
\end{tabular}  
\caption{Comparison of model accuracy, F1 score, and the inference cost for processing the CPT codes. ZR = Zero-shot Retriever; FR = Finetuned Retriever. Accuracy and F1 scores are reported on the validation set used in this study, while cost estimates reflect processing of the CPT codes. The second column indicates the type of context from the policy document supplied to each model, specifying whether the input was retrieved text passages or the full coverage document.}
\label{tab:cost_performance}  
\end{table*}

This section states both our results and our conclusions based on our observations of these results.
\subsection{Overall Performance}
% To facilitate rule tracing, we first evaluate the accuracy of decisions made using our approaches. Then we proceed with the interpretability.
To ensure that reviewers can trace how each rule is applied, we first assess the accuracy of the system’s outputs and then examine the interpretability of the explanations it provides.

Table \ref{tab:results} shows the average accuracy and F1 scores across the seven plan documents. The rule-based method using the finetuned retriever consistently outperforms the other two approaches. Finetuning improves accuracy by an average of 2.69\% and F1 score by 1.72\% over the zero-shot baseline. This improvement is largely due to the finetuned retriever's alignment with the language and structure of coverage policies. Trained on a large set of expert-annotated (CPT, subsection, relevance) pairs, it improves at pinpointing the policy passages most relevant to benefit considerations. Unlike standard semantic search, which may return thematically similar but irrelevant text, the finetuned cross-encoder model prioritizes concise policy fragments that matter for symbolic reasoning. Better passage selection directly enhances the quality of both attribute generation and rule creation in the symbolic phase, since rules are built from more accurate and relevant policy language. By capturing subtle cues and specific policy limitations, the finetuned retriever provides higher-quality inputs to the symbolic engine, resulting in more precise coverage determinations.

Overall, these results demonstrate that integrating expert-labeled supervision and contrastive ranking objectives into the retriever, together with symbolic reasoning, leads to measurable gains in performance.

\noindent \textbf{Interpretability.} One of our goals is to support interpretability, rather than relying solely on direct inference from the model (such as interpretation made purely through prompting). 
% For instance, as shown in Figure~\ref{fig:rule_tracing}, consider a procedure labeled \texttt{S9212} that is determined to be covered, under a specific policy, by the rule-based system. In this case, we can clearly observe that the rule \texttt{pregnancy\_maternity\_services} was triggered by the PyKnow engine. This rule is activated when the conditions \texttt{is\_pregnancy=True} and \texttt{is\_maternity=True} are met. Since these attributes accurately describe the procedure, the rule was applied, making the decision traceable and interpretable.
For instance, as illustrated in Figure \ref{fig:rule_tracing}, consider a procedure labeled \texttt{S9212}. Under the corresponding policy, the rule-based system identifies that the \texttt{pregnancy\_maternity\_services} rule is relevant. The PyKnow engine highlights this rule because its conditions, \texttt{is\_pregnancy=True} and \texttt{is\_maternity=True}, match the attributes associated with the procedure. This traceability allows a human reviewer to see exactly which factors contributed to the decision, providing transparency and context. The system thus serves as a support tool, helping humans make informed judgments. Ultimately, the final decision remains in the hands of the human reviewer. In contrast, direct prompting does not offer this level of traceability and is more prone to hallucinations, making it less reliable for such task.

\noindent \textbf{Cost Effectiveness.} Our rule-based approach is designed for scalability and cost efficiency in processing large volumes of clinical codes. Attribute generation is performed once for each CPT code; with over 11,000 CPT code, and even more when extending to other code sets like HCPCS, the ability to avoid repeated model inference becomes crucial. Rule generation is required only once per coverage policy, further reducing computational demands. During inference, our system relies solely on symbolic reasoning and minimal hardware; it does not require a GPU or LLM-based inference, making it both fast and inexpensive. As shown in Table~\ref{tab:cost_performance}, LLM-based methods such as GPT-5-mini, GPT-4.1, and o3 incur substantially higher costs, with GPT-5-mini costing approximately \$4,840 to process 11,000 CPT codes, compared to just \$22 for the finetuned rule-based system. Even with a one-time setup cost for training (i.e. \$2,680\footnote{Further details about the cost calculation are discussed in a later section.}), the rule-based approach remains highly cost-effective, especially as the number of codes grows. By eliminating the need for repeated LLM inference and leveraging symbolic reasoning, our method offers a robust and scalable solution for large-scale coverage adjudication across diverse clinical code sets.

\noindent \textbf{Performance vs Cost Tradeoff.} To further investigate cost effectiveness, we evaluated the use of retrieved texts as input for direct LLM inference in Table~\ref{tab:cost_performance}. Providing only the relevant passages instead of the entire coverage document significantly reduce the number of input tokens required for each inference, thereby lowering the overall cost. This approach allowed us to assess whether our retrieval-based method remains advantageous when combined with LLMs.

The cost rates for inference and setup vary notably across the different approaches. For LLM-based methods, API pricing is based on the number of input and output tokens processed: GPT-5-mini charges \$0.25 per million input tokens and \$2.00 per million output tokens, while both GPT-4.1 and o3 are priced at \$2.00 per million input tokens and \$8.00 per million output tokens. The rule-based system that uses finetuning involves a one-time training cost of approximately \$2,680, which was incurred by training on an external dataset for 48 hours using 8 × NVIDIA H100 GPUs (Azure H100 instances at 6.98/hour). Importantly, no GPUs are required for inference with the rule-based system, and the per-inference cost is extremely low, just \$2.50 per 1,000 CPT codes or \$22 for 11,000 codes.

LLM-based approaches such as GPT-5-mini, GPT-4.1, and o3 achieve the highest accuracy and F1 scores when using retrieved text as input, with average accuracy up to 0.94 and F1 scores up to 0.96. However, this performance comes at a substantial cost: processing 11,000 CPT codes costs \$4,840 for GPT-5-mini and \$9,680 for GPT-4.1 and o3. These methods also require a finetuned retriever for passage selection, adding a one-time setup cost of \$2,680. In contrast, the rule-based approaches deliver competitive performance (accuracy up to 0.87 and F1 up to 0.93) at a fraction of the cost ($\sim$\$2700 for 11,000 CPT codes). Notably, models that process entire documents are both less accurate (0.82 accuracy, 0.89 F1) and dramatically more expensive (\$38,720). Thus, while LLM-based methods provide slightly better accuracy with retrieved text, their inference costs scale rapidly with dataset size, making the rule-based approach far more cost-effective for large-scale adjudication tasks. Choosing between these methods involves balancing the goal of achieving optimal performance with the need to minimize operational costs. We plan to address this tradeoff in greater detail in future work.

\begin{figure*}[t]
    \centering  
    \includegraphics[width=1.0\textwidth]{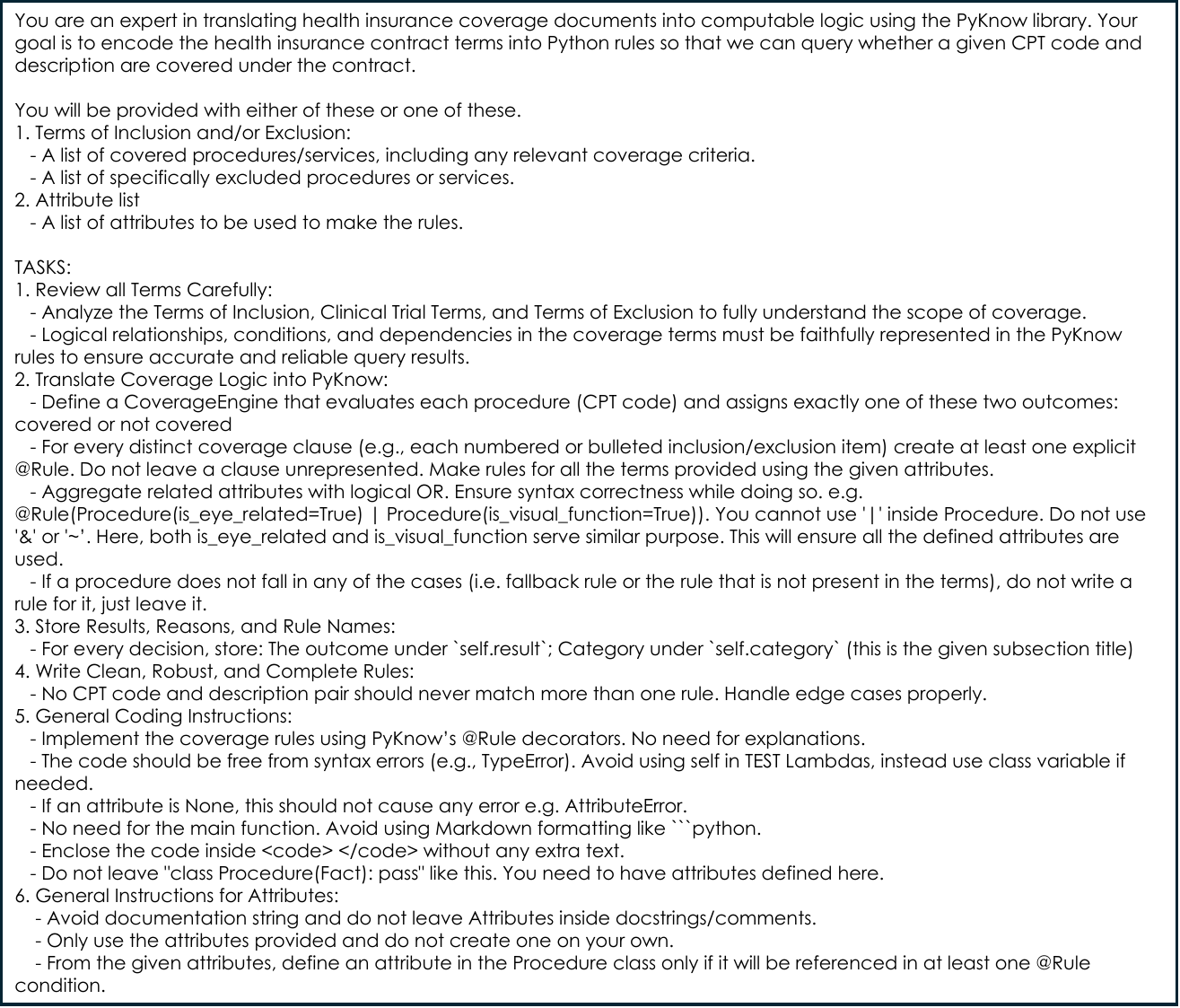}
    \caption{The prompt used for rule generation.}  
    \label{fig:prompt_rule}
\end{figure*}

\subsection{Where do the rules fail?} Our error analysis across all procedure codes and seven plans reveals that most rule failures can be attributed to two main factors. The first and most prevalent is that the correct attribute is sometimes not incorporated into the rule creation process. This accounts for 73.5\% of incorrect cases, where a rule is triggered but does not represent the correct logic. This often occurs when the attribute list is lengthy, leading the model to overlook or ``forget" attributes that appear later in the input sequence, a phenomenon also observed by~\cite{agrawal-etal-2024-evaluating, tao2025lost}. The second failure mode, making up the remaining 26.5\% of errors, involves instances where the model fails to generate a sufficient set of rules for certain samples, resulting in incomplete or missing logic and, thus, no rule being triggered for accurate coverage decisions. These findings underscore the importance of careful attribute selection and prompt engineering when constructing symbolic rules from complex policy language. Improving attribute coverage and accuracy is the most impactful path forward.

\section{Conclusion}

In this study, we introduced a framework designed to support human interpretability of coverage review by combining a coverage-aware retriever with a symbolic rule-based reasoning engine. Our findings suggest that finetuning the retriever with expert-labeled supervision and a contrastive ranking objective substantially improves its ability to surface the most relevant policy language compared to zero-shot approaches. This more accurate retrieval supports downstream steps such as attribute extraction and the identification of policy conditions that may be relevant for human reviewers. A key advantage of our approach is its efficiency: by narrowing the text that must be processed and reducing dependence on repeated LLM calls, the system provides significant cost savings while maintaining high performance. At the same time, limitations remain—for example, relevant attributes may be missed when long documents exceed input limits or when the underlying rule set does not fully capture policy nuances. Addressing these challenges will require continued refinement of context handling and rule construction. Overall, our results highlight the potential of neuro-symbolic methods to deliver scalable, interpretable, and cost-effective tools that support human reviewers by making policy logic more transparent and easier to trace.

\appendix

\section{Prompts}
\label{apd:prompts}

Figure~\ref{fig:prompt_rule} presents the structured prompts used during rule generation. 

\section{Acknowledgements}
We thank Ardavan Saeedi and the anonymous reviewers for their constructive feedback.

\bibliography{aaai2026}

@inproceedings{
    guha2023legalbench,
    title={LegalBench: A Collaboratively Built Benchmark for Measuring Legal Reasoning in Large Language Models},
    author={Neel Guha and Julian Nyarko and Daniel E. Ho and Christopher Re and Adam Chilton and Aditya Narayana and Alex Chohlas-Wood and Austin Peters and Brandon Waldon and Daniel Rockmore and Diego Zambrano and Dmitry Talisman and Enam Hoque and Faiz Surani and Frank Fagan and Galit Sarfaty and Gregory M. Dickinson and Haggai Porat and Jason Hegland and Jessica Wu and Joe Nudell and Joel Niklaus and John J Nay and Jonathan H. Choi and Kevin Tobia and Margaret Hagan and Megan Ma and Michael Livermore and Nikon Rasumov-Rahe and Nils Holzenberger and Noam Kolt and Peter Henderson and Sean Rehaag and Sharad Goel and Shang Gao and Spencer Williams and Sunny Gandhi and Tom Zur and Varun Iyer and Zehua Li},
    booktitle={Thirty-seventh Conference on Neural Information Processing Systems Datasets and Benchmarks Track},
    year={2023},
    url={https://openreview.net/forum?id=WqSPQFxFRC}
}

@inproceedings{yao-etal-2025-elevating,
    title = "Elevating Legal {LLM} Responses: Harnessing Trainable Logical Structures and Semantic Knowledge with Legal Reasoning",
    author = "Yao, Rujing  and
      Wu, Yang  and
      Wang, Chenghao  and
      Xiong, Jingwei  and
      Wang, Fang  and
      Liu, Xiaozhong",
    editor = "Chiruzzo, Luis  and
      Ritter, Alan  and
      Wang, Lu",
    booktitle = "Proceedings of the 2025 Conference of the Nations of the Americas Chapter of the Association for Computational Linguistics: Human Language Technologies (Volume 1: Long Papers)",
    month = apr,
    year = "2025",
    address = "Albuquerque, New Mexico",
    publisher = "Association for Computational Linguistics",
    url = "https://aclanthology.org/2025.naacl-long.290/",
    doi = "10.18653/v1/2025.naacl-long.290",
    pages = "5630--5642",
    ISBN = "979-8-89176-189-6"
}

@inproceedings{mishra-etal-2025-investigating,
    title = "Investigating the Shortcomings of {LLM}s in Step-by-Step Legal Reasoning",
    author = "Mishra, Venkatesh  and
      Pathiraja, Bimsara  and
      Parmar, Mihir  and
      Chidananda, Sat  and
      Srinivasa, Jayanth  and
      Liu, Gaowen  and
      Payani, Ali  and
      Baral, Chitta",
    editor = "Chiruzzo, Luis  and
      Ritter, Alan  and
      Wang, Lu",
    booktitle = "Findings of the Association for Computational Linguistics: NAACL 2025",
    month = apr,
    year = "2025",
    address = "Albuquerque, New Mexico",
    publisher = "Association for Computational Linguistics",
    url = "https://aclanthology.org/2025.findings-naacl.435/",
    doi = "10.18653/v1/2025.findings-naacl.435",
    pages = "7795--7826",
    ISBN = "979-8-89176-195-7"
}

@article{LegalfictionsDahletal2024,
    author = {Dahl, Matthew and Magesh, Varun and Suzgun, Mirac and Ho, Daniel E},
    title = {Large Legal Fictions: Profiling Legal Hallucinations in Large Language Models},
    journal = {Journal of Legal Analysis},
    volume = {16},
    number = {1},
    pages = {64-93},
    year = {2024},
    month = {06},
    issn = {2161-7201},
    doi = {10.1093/jla/laae003},
    url = {https://doi.org/10.1093/jla/laae003},
    eprint = {https://academic.oup.com/jla/article-pdf/16/1/64/58336922/laae003.pdf},
}

@inproceedings{Shengbin2024lawllm,
author = {Yue, Shengbin and Liu, Shujun and Zhou, Yuxuan and Shen, Chenchen and Wang, Siyuan and Xiao, Yao and Li, Bingxuan and Song, Yun and Shen, Xiaoyu and Chen, Wei and Huang, Xuanjing and Wei, Zhongyu},
title = {LawLLM: Intelligent Legal System with Legal Reasoning and Verifiable Retrieval},
year = {2024},
isbn = {978-981-97-5568-4},
publisher = {Springer-Verlag},
address = {Berlin, Heidelberg},
url = {https://doi.org/10.1007/978-981-97-5569-1_19},
doi = {10.1007/978-981-97-5569-1_19},
booktitle = {Database Systems for Advanced Applications: 29th International Conference, DASFAA 2024, Gifu, Japan, July 2–5, 2024, Proceedings, Part V},
pages = {304–321},
numpages = {18},
location = {Gifu, Japan}
}

@article{shen2025law,
  title={A Law Reasoning Benchmark for LLM with Tree-Organized Structures including Factum Probandum, Evidence and Experiences},
  author={Shen, Jiaxin and Xu, Jinan and Hu, Huiqi and Lin, Luyi and Zheng, Fei and Ma, Guoyang and Meng, Fandong and Zhou, Jie and Han, Wenjuan},
  journal={arXiv preprint arXiv:2503.00841},
  year={2025}
}

@article{shi2025legalreasoner,
  title={LegalReasoner: Step-wised Verification-Correction for Legal Judgment Reasoning},
  author={Shi, Weijie and Zhu, Han and Ji, Jiaming and Li, Mengze and Zhang, Jipeng and Zhang, Ruiyuan and Zhu, Jia and Xu, Jiajie and Han, Sirui and Guo, Yike},
  journal={arXiv preprint arXiv:2506.07443},
  year={2025}
}

@article{wei2022chain,
  title={Chain-of-thought prompting elicits reasoning in large language models},
  author={Wei, Jason and Wang, Xuezhi and Schuurmans, Dale and Bosma, Maarten and Xia, Fei and Chi, Ed and Le, Quoc V and Zhou, Denny and others},
  journal={Advances in neural information processing systems},
  volume={35},
  pages={24824--24837},
  year={2022}
}

@article{kant2025towards,
  title={Towards robust legal reasoning: Harnessing logical llms in law},
  author={Kant, Manuj and Nabi, Sareh and Kant, Manav and Scharrer, Roland and Ma, Megan and Nabi, Marzieh},
  journal={arXiv preprint arXiv:2502.17638},
  year={2025}
}

@inproceedings{ryu2025divide,
title={Divide and Translate: Compositional First-Order Logic Translation and Verification for Complex Logical Reasoning},
author={Hyun Ryu and Gyeongman Kim and Hyemin S. Lee and Eunho Yang},
booktitle={The Thirteenth International Conference on Learning Representations},
year={2025},
url={https://openreview.net/forum?id=09FiNmvNMw}
}

@inproceedings{pan-etal-2023-logic,
    title = "Logic-{LM}: Empowering Large Language Models with Symbolic Solvers for Faithful Logical Reasoning",
    author = "Pan, Liangming  and
      Albalak, Alon  and
      Wang, Xinyi  and
      Wang, William",
    editor = "Bouamor, Houda  and
      Pino, Juan  and
      Bali, Kalika",
    booktitle = "Findings of the Association for Computational Linguistics: EMNLP 2023",
    month = dec,
    year = "2023",
    address = "Singapore",
    publisher = "Association for Computational Linguistics",
    url = "https://aclanthology.org/2023.findings-emnlp.248/",
    doi = "10.18653/v1/2023.findings-emnlp.248",
    pages = "3806--3824"
}

@inproceedings{yang-etal-2024-harnessing,
    title = "Harnessing the Power of Large Language Models for Natural Language to First-Order Logic Translation",
    author = "Yang, Yuan  and
      Xiong, Siheng  and
      Payani, Ali  and
      Shareghi, Ehsan  and
      Fekri, Faramarz",
    editor = "Ku, Lun-Wei  and
      Martins, Andre  and
      Srikumar, Vivek",
    booktitle = "Proceedings of the 62nd Annual Meeting of the Association for Computational Linguistics (Volume 1: Long Papers)",
    month = aug,
    year = "2024",
    address = "Bangkok, Thailand",
    publisher = "Association for Computational Linguistics",
    url = "https://aclanthology.org/2024.acl-long.375/",
    doi = "10.18653/v1/2024.acl-long.375",
    pages = "6942--6959"
}

@inproceedings{vakharia2024proslm,
  title={Proslm: A prolog synergized language model for explainable domain specific knowledge based question answering},
  author={Vakharia, Priyesh and Kufeldt, Abigail and Meyers, Max and Lane, Ian and Gilpin, Leilani H},
  booktitle={International Conference on Neural-Symbolic Learning and Reasoning},
  pages={291--304},
  year={2024},
  organization={Springer}
}

@article{tan2024thought,
  title={Thought-Like-Pro: Enhancing Reasoning of Large Language Models through Self-Driven Prolog-based Chain-of-Thought},
  author={Tan, Xiaoyu and Deng, Yongxin and Qiu, Xihe and Xu, Weidi and Qu, Chao and Chu, Wei and Xu, Yinghui and Qi, Yuan},
  journal={arXiv preprint arXiv:2407.14562},
  year={2024}
}

@inproceedings{xu-etal-2024-faithful,
    title = "Faithful Logical Reasoning via Symbolic Chain-of-Thought",
    author = "Xu, Jundong  and
      Fei, Hao  and
      Pan, Liangming  and
      Liu, Qian  and
      Lee, Mong-Li  and
      Hsu, Wynne",
    editor = "Ku, Lun-Wei  and
      Martins, Andre  and
      Srikumar, Vivek",
    booktitle = "Proceedings of the 62nd Annual Meeting of the Association for Computational Linguistics (Volume 1: Long Papers)",
    month = aug,
    year = "2024",
    address = "Bangkok, Thailand",
    publisher = "Association for Computational Linguistics",
    url = "https://aclanthology.org/2024.acl-long.720/",
    doi = "10.18653/v1/2024.acl-long.720",
    pages = "13326--13365"
}

@inproceedings{toroghi-etal-2024-verifiable,
    title = "Verifiable, Debuggable, and Repairable Commonsense Logical Reasoning via {LLM}-based Theory Resolution",
    author = "Toroghi, Armin  and
      Guo, Willis  and
      Pesaranghader, Ali  and
      Sanner, Scott",
    editor = "Al-Onaizan, Yaser  and
      Bansal, Mohit  and
      Chen, Yun-Nung",
    booktitle = "Proceedings of the 2024 Conference on Empirical Methods in Natural Language Processing",
    month = nov,
    year = "2024",
    address = "Miami, Florida, USA",
    publisher = "Association for Computational Linguistics",
    url = "https://aclanthology.org/2024.emnlp-main.379/",
    doi = "10.18653/v1/2024.emnlp-main.379",
    pages = "6634--6652"
}

@article{cummins2025computable,
  title={Computable contracts for insurance: Establishing an insurance-specific controlled natural language-insurle},
  author={Cummins, J and D{\'a}vila, J and Kowalski, R and Ovenden, D},
  journal={Accessed: Feb},
  volume={11},
  pages={2025},
  year={2025}
}

@article{garrido2025gofai,
  title={GOFAI meets Generative AI: Development of Expert Systems by means of Large Language Models},
  author={Garrido-Merch{\'a}n, Eduardo C and Puente, Cristina},
  journal={arXiv preprint arXiv:2507.13550},
  year={2025}
}

@article{li2024mediq,
  title={Mediq: Question-asking llms and a benchmark for reliable interactive clinical reasoning},
  author={Li, Stella and Balachandran, Vidhisha and Feng, Shangbin and Ilgen, Jonathan and Pierson, Emma and Koh, Pang Wei W and Tsvetkov, Yulia},
  journal={Advances in Neural Information Processing Systems},
  volume={37},
  pages={28858--28888},
  year={2024}
}

@article{oord2018representation,
  title={Representation learning with contrastive predictive coding},
  author={Oord, Aaron van den and Li, Yazhe and Vinyals, Oriol},
  journal={arXiv preprint arXiv:1807.03748},
  year={2018}
}

@inproceedings{agrawal-etal-2024-evaluating,
    title = "Evaluating Multilingual Long-Context Models for Retrieval and Reasoning",
    author = "Agrawal, Ameeta  and
      Dang, Andy  and
      Bagheri Nezhad, Sina  and
      Pokharel, Rhitabrat  and
      Scheinberg, Russell",
    editor = {S{\"a}lev{\"a}, Jonne  and
      Owodunni, Abraham},
    booktitle = "Proceedings of the Fourth Workshop on Multilingual Representation Learning (MRL 2024)",
    month = nov,
    year = "2024",
    address = "Miami, Florida, USA",
    publisher = "Association for Computational Linguistics",
    url = "https://aclanthology.org/2024.mrl-1.18/",
    doi = "10.18653/v1/2024.mrl-1.18",
    pages = "216--231"
}

@article{tao2025lost,
  title={"Lost-in-the-Later": Framework for Quantifying Contextual Grounding in Large Language Models},
  author={Tao, Yufei and Hiatt, Adam and Seetharaman, Rahul and Agrawal, Ameeta},
  journal={arXiv preprint arXiv:2507.05424},
  year={2025}
}

@article{nezhad2025symcode,
  title={SymCode: A Neurosymbolic Approach to Mathematical Reasoning via Verifiable Code Generation},
  author={Nezhad, Sina Bagheri and Li, Yao and Agrawal, Ameeta},
  journal={arXiv preprint arXiv:2510.25975},
  year={2025}
}

@InProceedings{pmlr-v284-nezhad25a,
  title = 	 {Enhancing Large Language Models with Neurosymbolic Reasoning for Multilingual Tasks},
  author =       {Nezhad, Sina Bagheri and Agrawal, Ameeta},
  booktitle = 	 {Proceedings of The 19th International Conference on Neurosymbolic Learning and Reasoning},
  pages = 	 {1059--1076},
  year = 	 {2025},
  editor = 	 {H. Gilpin, Leilani and Giunchiglia, Eleonora and Hitzler, Pascal and van Krieken, Emile},
  volume = 	 {284},
  series = 	 {Proceedings of Machine Learning Research},
  month = 	 {08--10 Sep},
  publisher =    {PMLR},
  url = 	 {https://proceedings.mlr.press/v284/nezhad25a.html}
}

\end{document}